\documentclass{article}
\usepackage{ijcai26}

\usepackage{times}
\usepackage{soul}
\usepackage{url}
\usepackage[hidelinks]{hyperref}
\usepackage[utf8]{inputenc}
\usepackage[small]{caption}
\usepackage{graphicx}
\usepackage{amsmath}
\usepackage{amsthm}
\usepackage{booktabs}
\usepackage{algorithm}
\usepackage{algorithmic}
\usepackage[switch]{lineno}
\usepackage{comment} 
\usepackage{enumitem}
\usepackage{siunitx}
\usepackage{todonotes}
\usepackage{xspace}

\usepackage{subcaption}
 
\newcommand{\epl}{$p^*_L$\xspace}
\newcommand{\eph}{$p^*_H$\xspace}
\newcommand{\ep}{$p^*$\xspace}

\newcommand{\epinit}{$p_0$\xspace}
\newcommand{\sol}{$\mathit{SOL}$\xspace}
\newcommand{\sols}{$\mathit{SOL}$s\xspace}
\newcommand{\cpo}{CPO\xspace}
\newcommand{\clingo}{clingo\xspace}

\title{Learning to Solve and Optimize by Evolving Code}
\author{
{Veronika Semmelrock}\thanks{The first three authors contributed equally and are listed in alphabetical order. The same holds for the last three authors.}$^{1}$
\and
{Benedetta Strizzolo}$^{*2}$ \and
{Francesco Zuccato}$^{*1}$ \and \\
{Gerhard Friedrich}$^{*1}$ \and
{Patrick Rodler}$^{*1}$ \And
{Konstantin Schekotihin}$^{*1}$\\
\affiliations
$^1$University of Klagenfurt, Austria\\
$^2$University of Udine, Italy\\
\emails
\{firstname.lastname\}@aau.at,
strizzolo.benedetta@spes.uniud.it
}

\begin{document}

\maketitle

\begin{abstract}
Combinatorial and optimization problems are fundamental to many industrial AI applications.
Solving large-scale real-world instances of such problems typically requires careful problem formalization, specialized solvers, and expert-designed heuristics.
Thus, experts need to specify not only \textit{what} solutions are, but also \textit{how} they are derived.

By introducing the tool \textsc{CheckMate}, we show that algorithm generation via code evolution represents a paradigm shift by eliminating the need to formulate the \textit{how}.
\textsc{CheckMate} solely relies on the \textit{what}.
% Specifically, a natural language description guides the programs' evolutionary process, and the formal specification enables systematic performance evaluation over a sample of problem instances.
Specifically, a formal specification ensures solutions' correctness and enables systematic performance evaluation of the generated programs, while a natural language description guides the evolutionary process.

The effectiveness of our method is demonstrated on selected problems from two industrial domains: configuration and scheduling.
In all cases, the evolved algorithms consistently outperform state-of-the-art solvers.
This underscores the potential of formal methods in guiding code evolution for automatically solving complex real-world problems.
\end{abstract}

%%%%%%%%%%%%%%%%%%%%%%%%%%%%%%%%%%%%%%%%%%%%%%%%%%%%%%%%%%%%%%%%%%%%%%%%%%%%
\section{Introduction}
\label{sec:intro}

Combinatorial and optimization problems are central to many industrial AI applications, including the automated engineering of technical systems (e.g., configuring and planning the composition of large electronic systems) \cite{FalknerFHSS16} and the automation of process planning and scheduling \cite{da2022industrial}. A long-standing vision in AI is that domain experts should specify \emph{what} constitutes a correct solution, while the computer should determine \emph{how} to obtain it efficiently \cite{Freuder18}.

In practice, however, applying state-of-the-art solvers from constraint \cite{laborie2018ibm}, logic \cite{KaufmannLPS16}, or mathematical programming \cite{perron_et_al:LIPIcs.CP.2023.3} to large real-world problem instances still demands substantial expert intervention beyond the \emph{what}. Effective deployment often requires redesigning problem specifications \cite{DodaroGGMMMS24}, crafting problem-specific heuristics \cite{Comploi-TaupeFS23}, decomposing the problem into manageable subproblems \cite{ElKholanyGS25}, or implementing tailored local-search procedures \cite{SanghikianMSM26}.

To address these challenges, we build on recent advances in program generation via code evolution \cite{novikov2025alphaevolve}. Our approach integrates a new \textsc{CheckMate} component into the OpenEvolve \cite{openevolve} controller loop. Given a natural-language description and a formal specification of \emph{what} solutions are, along with a training set of representative instances, OpenEvolve+\textsc{CheckMate} automatically synthesizes a problem-specific solver—implemented as a Python program—without requiring any formulation of \emph{how} to solve the problem.

We investigate the following research questions:
\begin{enumerate}[topsep=0pt,noitemsep,label=\textit{RQ\arabic*}, ref=RQ\arabic*, labelwidth=3em, labelsep=0.5em, leftmargin=!]
    \item \label{rq1:solving} Can 
    %contemporary 
    OpenEvolve+\textsc{CheckMate} solve hard real-world \emph{(a)}~combinatorial and \emph{(b)}~optimization problems, such as configuration and scheduling?
    \item \label{rq2:vs-sota} How does the performance of the generated programs compare to state-of-the-art solvers?
    \item \label{rq3:scalability} How well do these evolved programs scale?
\end{enumerate}

Our evaluation targets configuration problems from automated engineering and an industrial scheduling problem. We analyze two configuration tasks from Siemens that capture pivotal real-world challenges: one stresses solver scalability with respect to solution size \cite{semmelrock26}, and the other examines solver behavior when several difficult problems are combined \cite{Gebser2015SolvingCC}. In addition, we evaluate a real-world scheduling problem from metalworking at voestalpine \cite{zuccato2025energy}.

Across all tasks, OpenEvolve+\textsc{CheckMate} generated problem-specific code that outperformed the respective state-of-the-art solvers by orders of magnitude on large or hard instances. Thus, our approach extends to solving problems where leading solvers cannot deliver solutions. 

%%%%%%%%%%%%%%%%%%%%%%%%%%%%%%%%%%%%%%%%%%%%%%%%%%%%%%%%%%%%%%%%%%%%%%%%%%%%
\section{Preliminaries}
\label{sec:preliminaries}
Since our contribution integrates with OpenEvolve \cite{openevolve}, an open-source implementation of AlphaEvolve \cite{novikov2025alphaevolve}, we provide a brief overview of this framework.
OpenEvolve is a code optimization framework that automatically generates programs capable of solving specified problems by leveraging evolutionary computation and the capabilities of Large Language Models (LLMs) in program synthesis.
This framework operates through an asynchronous pipeline consisting of four core modules: \textit{(i)}~a \emph{prompt sampler} that generates prompts using previously evolved programs and their evaluation scores, \textit{(ii)}~an \emph{LLM ensemble}, i.e., a set of LLMs that process prompts and produce full code rewrites or targeted edits, \textit{(iii)}~an \emph{evaluator pool} that scores and ranks generated programs, and \textit{(iv)}~a \emph{program database} that stores these programs and their scores in a structured grid~\cite{mouret2015illuminatingsearchspacesmapping}, organizing them according to diverse user-defined criteria, such as code complexity and correctness. 
Note that the evaluator pool, at its core, is an evaluation function that the user must implement.
An island-based genetic algorithm \cite{DBLP:phd/us/Tanese89} manages selection, migration, and evolution across iterations.

Adopting OpenEvolve's terminology, we denote an \textit{artifact} as a textual error-feedback provided to the LLM, and the \textit{combined score} $z_j\in[0,1]$ as a numerical score assigned to program $p_j$. %, used for ranking. 
Intuitively, the higher the combined score, the higher the likelihood for the program to be selected by the prompt sampler and to guide further evolution. 

%%%%%%%%%%%%%%%%%%%%%%%%%%%%%%%%%%%%%%%%%%%%%%%%%%%%%%%%%%%%%%%%%%%%%%%%%%%%
\section{Approach}
\label{sec:approach}

\begin{figure*}[t]  % * for double-column   
    \centering
    \includegraphics[width=0.8\textwidth]{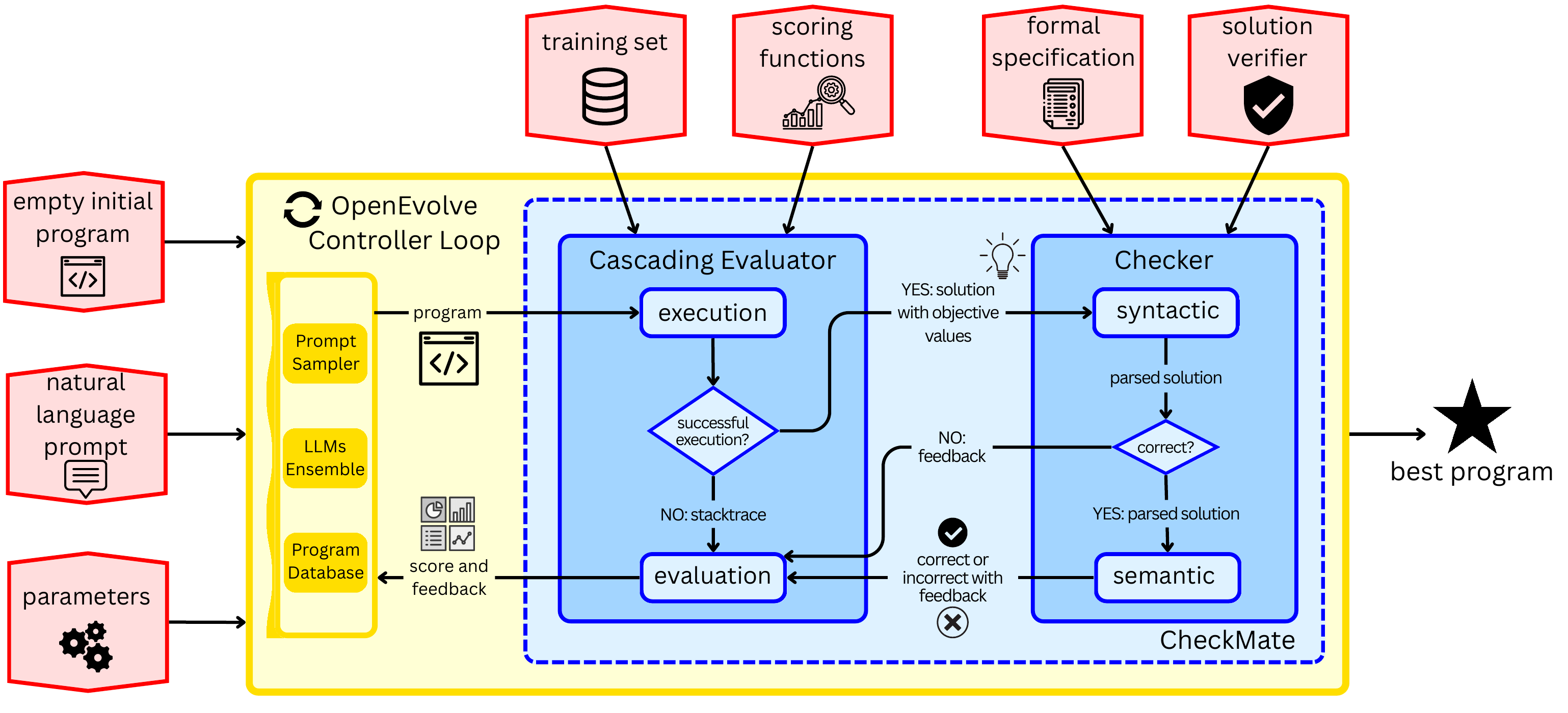} %columnwidth
    \caption{System overview of the proposed evolution framework}
    \label{fig:img/system_overview.png}
\end{figure*}

%Generalizing the OpenEvolve pipeline described in Sec.~\ref{sec:preliminaries}, we introduce \textsc{CheckMate}, which provides a problem-independent implementation of the \emph{evaluator pool} that employs state-of-the-art solvers as verifiers to check the correctness of candidate solutions.
Extending OpenEvolve's pipeline (Sec.~\ref{sec:preliminaries}), our approach \textsc{CheckMate} provides a general, problem-independent implementation of the \emph{evaluator pool} that employs state-of-the-art solvers as verifiers to check the correctness of programs' produced problem solutions. %program outputs.
Fig.~\ref{fig:img/system_overview.png} illustrates the integrated framework, with our contributions highlighted in the dashed blue box.
OpenEvolve+\textsc{CheckMate} expects the following inputs: \textit{(i)}~a natural language description $D$ of the problem (prompt), \textit{(ii)}~a formal specification $\mathit{F}$ defining valid problem solutions, \textit{(iii)}~a solution verifier $V$, \textit{(iv)}~a set of training problem instances $I$ (for which valid problem solutions are assumed to exist)%\footnote{$I$ only consists of program inputs, no outputs are provided.}
, \textit{(v)}~a set $S$ of scoring functions for program evaluation, \textit{(vi)}~a (possibly empty) initial program \epinit, and \textit{(vii)}~a set of evolution and LLM hyperparameters $\Lambda$.

Given the inputs, OpenEvolve+\textsc{CheckMate}, denoted as the function generator $g$, produces a program \ep. 
Formally:
\[ g : (\mathit{D},\mathit{F},\mathit{V},\mathit{I},\mathit{S},\mathit{p_0},\Lambda) \mapsto p^*  \]
The generation of \ep involves $N$ training iterations, with $N$ defined in $\Lambda$.
At each iteration $1 \le j \le N$, the LLM ensemble non-deterministically generates an intermediate program $p_j$, aiming to improve the programs $p_0, \ldots, p_{j-1}$.
\textsc{CheckMate} evaluates $p_j$ by producing the combined score $z_j$.
After $N$ iterations, the $p_j$ achieving the highest $z_j$ will then correspond to the final best program \ep.

At the core of \textsc{CheckMate}, the solution verifier $V$ is used to check the correctness of program outputs.
Specifically, given a program $p_j$, which (possibly) generates a candidate solution $c_{ji}$ for a problem instance $i$, $V$ outputs \emph{(i)}~a $\mathit{verdict}\in\{ \mathit{correct}, \mathit{incorrect} \}$ indicating whether $c_{ji}$ is an (in)correct solution for $\mathit{i}$, and, \emph{(b)}~in case of $\mathit{incorrect}$, some $\mathit{feedback}$ detailing which parts of the formal specification $F$ are violated by $c_{ji}$.
Formally:
\[ p_j : \mathit{i} \mapsto \mathit{c_{ji}} \qquad V : (F, i, c_{ji}) \mapsto \langle \mathit{verdict}, \mathit{feedback} \rangle \]
\textsc{CheckMate}'s overall program evaluation process, employing $V$, is detailed next.

\subsection{Program Execution and Evaluation}
\label{sec:heur-eval}
In each training iteration $j \le N$, in order to evaluate the generated program $p_j$, $\textsc{CheckMate}$:
\begin{enumerate}[topsep=0pt,noitemsep, label=(\arabic*)]
    \item \label{enum:checking:exec} Executes $p_j$ on a training instance $i \in I$: if $p_j$ fails, go to \ref{enum:checking:eval} and forward $\langle \mathit{incorrect}, \mathit{feedback} \rangle $, where the feedback details the failure reason, e.g., an error stack trace; otherwise, go to \ref{enum:checking:synt} and forward candidate solution $c_{ji}$;
    \item \label{enum:checking:synt} Syntactically checks $c_{ji}$: if the check fails, go to \ref{enum:checking:eval} and forward $\langle \mathit{incorrect}, \mathit{feedback} \rangle$, where the feedback includes, e.g., syntactic errors in $c_{ji}$; else, go to \ref{enum:checking:sem} and forward $c_{ji}$ parsed into the verifier-specific format;
    \item \label{enum:checking:sem} Checks the correctness of $c_{ji}$ using the verifier~$V$: if the check returns positively, go to \ref{enum:checking:eval} and forward $\langle \mathit{correct}, \emptyset \rangle$; else, go to \ref{enum:checking:eval} and forward $\langle \mathit{incorrect}, \mathit{feedback} \rangle$, where feedback includes, e.g., violated constraints in $F$;
    \item \label{enum:checking:eval} Stores $\mathit{feedback}$ and evaluates $p_j$'s performance on instance $i$ using the scoring functions $S$, based on the verdict ($\mathit{correct}$/$\mathit{incorrect}$) as well as statistics such as runtime, consumed memory, and resulting objective values.
\end{enumerate}
This process is repeated for all training instances $i \in I$, or until the early stopping is triggered due to $k$ consecutive $\mathit{incorrect}$ verdicts.\footnote{On early stopping, unevaluated instances are deemed \textit{incorrect}.}

Finally, \textsc{CheckMate} \emph{(i)}~aggregates the instance-level scores from Step (4) to determine $z_j$, the combined score of $p_j$, and \emph{(ii)}~returns $z_j$ with the stored feedback for all instances (i.e., artifacts) to OpenEvolve's main controller loop.

\subsection{Textual Feedback}
\label{sec:textual-feedback}
The artifacts (cf.\ Sec.~\ref{sec:preliminaries}) are used to provide textual feedback directly to the LLM.
They depend on the underlying failure type, of which we distinguish five kinds: \emph{(i)}~\textit{execution} exceptions, such as compilation or runtime errors, \emph{(ii)}~\textit{intentional} exceptions, raised by the evolved program itself due to a precondition or invariant violation, \emph{(iii)}~\textit{exceeded resource} errors, whenever (user-defined) time or memory limits are reached, \emph{(iv)}~\textit{syntactic} errors, e.g., an incomplete candidate solution, and \emph{(v)}~\textit{semantic} errors, if the verdict of $V$ for $c_{ji}$ equals $\mathit{incorrect}$. 

Each artifact comprises the failed instance, the failure type, and a suggestion for repairing the failure. The suggestion proposes an improvement to the program in natural language, depending on the failure type. It can include the stack trace, the error message, the position of syntactic errors, or the set of violated constraints or logical sentences.
The artifact is stored alongside program $p_j$ and is injected into the prompt whenever $p_j$ is selected in subsequent iterations.

\paragraph{Failure Protocol: Self-refined feedback on program-level.}
\label{sec:failure-prot}
To support the evolution process when programs do not produce candidate solutions, we introduce the Failure Protocol as part of the textual feedback mechanism. By instructing the LLM via the prompt to implement this protocol, we enable a program-level self-refinement loop.
Whenever the program expects that the ongoing candidate solution generation will fail, it should raise one of the following exceptions, indicating that it believes \emph{(i)}~the instance has no correct solution, \emph{(ii)}~the instance has a correct solution but the implemented search strategy will not find it, or \emph{(iii)}~it cannot recover from wrong decisions.

Exceptions raised in accordance with the Failure Protocol are referred to as \textit{intentional} exceptions. They include additional information that the program considers useful feedback for evolution. During execution, \textsc{CheckMate} intercepts intentional exceptions and parses them into artifacts, including the suggestion that the error indeed lies within the program, because all instances are assumed to be satisfiable. 
%Note, instances of many combinatorial optimization problems can trivially be satisfied given a sufficiently large number of constants, e.g., an unlimited number of components in configuration or a very long time interval in scheduling.

\subsection{Inputs}
\label{sec:inputs}
In this section, we present the inputs to the code evolution framework that are either novel or central to our approach.
\paragraph{Formal specification and solution verifier.}
The formal specification $F$ is used by the verifier $V$ to check if a candidate solution $c_{ij}$ yielded by the program $p_j$ is correct for the problem instance $i$. 
It can be provided, e.g., in the form of constraints or logical sentences depending on the selected $V$, such as \clingo~\cite{potasscoguide}, or CP Optimizer~\cite{laborie2018ibm}.
Since $F$ is used solely to check,
there is no need to tune it for solving, e.g., by employing guessing or symmetry-breaking techniques. %for tuning in order to prune the search space.
\paragraph{Prompt.}
The natural language description $D$ is a solver-agnostic specification designed to be independent from the format of the solution verifier. It is structured as a domain-specific prompt which contains:
\textit{(i)}~the LLM identity, \textit{(ii)}~the task to be performed, \textit{(iii)}~the problem description including context, core entities, and constraints that must be satisfied, and \textit{(iv)}~instructions on expected input/output formats and algorithm requirements.
This structured approach provides the LLM with the necessary context for generating programs.

\paragraph{Scoring functions.} 
The scoring functions constitute one of the most fundamental components of the learning process, as they affect how programs are ranked, selected, and evolved across iterations.
As introduced in Sec.~\ref{sec:heur-eval}, they are employed to derive $z_j$, the combined score of each program $p_j$.
Specifically, $z_j$ is built upon the correctness $z_{j}^{\mathit{c}}$ and the quality-efficiency $z_{j}^{\mathit{qe}}$ trade-off scores.
The correctness score builds upon the number of training instances $i \in I$ solved, while the quality-efficiency score combines the quality score $z_{j}^{q}$ and the efficiency score $z_{j}^{e}$.
The quality score reflects the optimization of the objective values, whereas the efficiency score combines the runtime and memory consumption.

In particular, \textsc{CheckMate} \emph{(i)}~normalizes the raw statistics into scores in $[0,1]$;\footnote{
Normalization must ensure the maximization of each score. For decision problems, we assign a trivial quality score of $1$ to any correct solution, whereas for optimization problems, we require lower and upper bounds for each objective of every instance for scoring.
%Normalization must ensure maximization of each score. Problems where generating any correct solution is the goal yield trivial quality scores of $1$, whereas optimization problems require lower and upper bounds for each objective of every instance for scoring.
}
\emph{(ii)}~aggregates the instance-level scores into the overall correctness $z_{j}^{c}$, quality $z_{j}^{q}$, and efficiency $z_{j}^{e}$ scores;
\emph{(iii)}~combines the quality and efficiency scores into the composite quality-efficiency tradeoff score $z_{j}^{\mathit{qe}}$;
and \emph{(iv)}~combines $z_{j}^{c}$, $z_{j}^{\mathit{qe}}$ into the combined score $z_{j}$.
To address these tasks, we employed an exponential decay function for Step (i), the arithmetic mean for (ii), and combinations of harmonic mean, product, and Prioritized Ordered Weighted Average ($\mathit{POWA}$) \cite{yager2009prioritized} for (iii) and (iv).

\paragraph{Training set.}\label{sec:trainingset}
The training set should include representative instances of various difficulty levels.
Since the initial program \epinit is empty, including sufficiently easy instances is highly recommended to enable the evolutionary process to get started. In early iterations, the main challenge is to solve at least some instances. Once this occurs, the scoring function becomes informative, yielding non-zero correctness values and therefore a useful combined score, enabling the program to evolve effectively. As training progresses, it appears that instances of moderate difficulty promote generalization, while hard ones support scaling and efficiency improvements.

%%%%%%%%%%%%%%%%%%%%%%%%%%%%%%%%%%%%%%%%%%%%%%%%%%%%%%%%%%%%%%%%%%%%%%%%%%%%
\section{Case Studies}
\label{sec:casestudies}
In this work, we demonstrate the effectiveness of our approach on the three case studies motivated in Sec.~\ref{sec:intro}. 
This section provides a high-level overview of these case studies.
% The natural language description and the formal specification are provided in \cite{checkmate2026repo}.

\subsection{House Configuration Problem}
The technology-independent House Configuration Problem (HCP) \cite{FleischanderlFHSS98,Rya11} was introduced by Siemens and abstracts electronic modules, frames, and racks into things, cabinets, and rooms.
Specifically, persons own multiple things, whereas each thing belongs to exactly one person.
The task is to assign things to cabinets and cabinets to rooms while satisfying capacity, ownership, and ordering constraints (the latter introduced in \cite{semmelrock26}).

\subsection{Combined Configuration Problem}
The Combined Configuration Problem (CCP) \cite{Gebser2015SolvingCC} is a combinatorial benchmark motivated by industrial product configuration tasks at Siemens, such as the configuration of railway control and safety systems \cite{FalknerFHSS16}. The problem is defined via a directed acyclic graph with vertices, paths, bins, colors, areas, and border elements. 
The CCP integrates multiple interacting subproblems that must be solved simultaneously: \emph{(P1)} vertex coloring, \emph{(P2)} bin packing, \emph{(P3)} partitioning into disjoint paths, \emph{(P4)} matching border elements to areas, and \emph{(P5)} ensuring the connectedness of color-induced subgraphs. 
The combination of these subproblems makes the CCP very challenging for solvers.

\subsection{Energy-Aware Double-Flexible Job-Shop}
\label{sec:edfjsp}
The Energy-Aware Double-Flexible Job Shop Problem (E-DFJSP) scheduling problem \cite{gong2018new}  
exemplifies a real-world production process at voestalpine\footnote{\url{https://www.voestalpine.com/}}, where jobs consist of multiple metal-cutting operations.
Each cut operation requires selecting an eligible machine and its corresponding parameters, from a $k \times k$ grid. Between cuts, workers perform setup and transport operations on the machines.
A correct schedule assigns the operations to the corresponding machines and workers while satisfying all constraints, such as preventing resource overlaps.
Schedules are evaluated according to lexicographically prioritized objectives to be minimized: 
\emph{(1)} job tardiness $\mathit{Tard}$, %, i.e., the lateness of the jobs with respect to their due date, 
\emph{(2)} total energy consumption $\mathit{TEC}$---including auxiliary, idle, and processing energy---, and 
\emph{(3)} makespan $C_\mathit{max}$. %, i.e., the total duration of the schedule.
Note that, as the machine parameters influence both processing time and energy consumption, optimizing such a problem is a highly complex task.

We map the objective values to the quality score $z_j^q$ as follows.
First, $\mathit{Tard}$, $\mathit{TEC}$, and $C_\mathit{max}$ are normalized into $[0,1]$ using the corresponding theoretical lower and upper bounds.
Next, an exponential decay function is applied to compute the scores $z_j^{\mathit{Tard}}$, $z_j^{\mathit{TEC}}$, and $z_j^{C_\mathit{max}}$, which are subsequently combined into $z_j^q = \mathit{POWA}(z_j^{\mathit{Tard}},$  $z_j^{\mathit{TEC}}, z_j^{C_\mathit{max}})$.

%%%%%%%%%%%%%%%%%%%%%%%%%%%%%%%%%%%%%%%%%%%%%%%%%%%%%%%%%%%%%%%%%%%%%%%%%%%%

\section{Evaluation}
\label{sec:eval}
In this section, we evaluate the proposed approach experimentally. \cite{checkmate2026repo} provides datasets, evolved programs, and results.
\subsection{Datasets}
\label{sec:datasets}
For each case study, we use separate training and test sets, all comprising easy, moderate, and hard instances.

For the CCP, we reuse the dataset of \citeauthor{Gebser2015SolvingCC} (\citeyear{Gebser2015SolvingCC}), which comprises $99$ instances: $30$ easy, $33$ moderate, and $36$ hard, which, for brevity, we will denote by $30/33/36$. 
For the \emph{training set}, we select $20$ instances, split as $3/7/10$, as such a split was employed to define the CCP's competition instances for the 6th ASP Competition \cite{gebser2017sixth}. %\footnote{\url{https://aspcomp2015.dibris.unige.it}}
Moreover, we generate and include $5$ more \textit{very} easy instances (cf.\ Sec.~\ref{sec:trainingset}) by reducing the number of components in the easy ones. Hence, the CCP's training set includes $25$ instances ($8/7/10$), while the \emph{test set} comprises the $79$ remaining ASP competition instances ($27/26/26$).% (incl.\ remaining ASP competition instances).
 
For the HCP and the E-DFJSP, we generate both training and test sets, ensuring that all instances are satisfiable by design. %, a property that is not trivial to achieve for the HCP instance.
The \emph{training sets} are composed of $15$ instances, split as $5/5/5$.
For the HCP, the easy instances consist of up to $5$ persons ($\mathord{\sim} 50$ things), the moderate instances 
%consist 
of up to $50$ persons ($\mathord{\sim} 500$ things), and the hard instances of up to $500$ persons ($\mathord{\sim} 5\,000$ things). The hardness assessments reflect the results of \cite{semmelrock26} on \clingo's performance.
Likewise, for the E-DFJSP, the easy instances involve up to $10$ jobs ($\mathord{\sim} 60$ operations), the moderate up to $50$ jobs ($\mathord{\sim} 300$ operations), and the hard up to $500$ jobs ($\mathord{\sim}3\,000$ operations).
This is in accordance with \cite{schlenkrich2023solving} where scheduling problems were classified as \textit{large-scale} if they comprise at least $1\,000$ operations.

We designed the \emph{test sets} to contain an overproportional number of hard instances ranging from large to \emph{very} large size, in accordance with research question \ref{rq3:scalability}, i.e., investigating the scalability of the evolved code.
Specifically, we used a split of $6/6/24$, where the $24$ hard cases comprise up to $3\,000$ persons ($\mathord{\sim} {30\,000}$ things) for the HCP and up to $1\,000$ jobs ($\mathord{\sim} 6\,000$ operations) for the E-DFJSP, thus going significantly beyond what is commonly already considered hard.

\subsection{Experiments}
\label{sec:exp}
Each experiment was conducted on a machine equipped with an AMD EPYC 7H12 64-core Processor, with RAM usage restricted to 32~GB.
Timeouts for evolved programs and solvers to find a solution for any problem were set to 600\,s.
\paragraph{Training.}
Since OpenEvolve+\textsc{CheckMate} is inherently non-deterministic, each evolutionary training run can produce a different best program \ep.
To account for this variability, we performed four independent training runs for each case study (Sec.~\ref{sec:casestudies}), yielding four corresponding best programs \ep.

Each problem has a specific prompt $D$ and formal specification $F$. 
Exclusively for CCP, due to its hardness, \emph{(i)}~$D$ includes the Failure Protocol (Sec.~\ref{sec:failure-prot}), and \emph{(ii)}~$F$ is exploited to pinpoint violated constraints returned by the verifier whenever its verdict for a candidate solution equals \emph{incorrect}.

While most parameters were kept at OpenEvolve's default values, the following problem-specific settings were used: \textit{(i)}~the number of evolutionary iterations ($50$ for HCP, $75$ for CCP, $100$ for E-DFJSP, or, for brevity $50/75/100$), \textit{(ii)}~the number of islands ($3/5/5$), and \textit{(iii)}~the migration interval ($10/10/15$) to promote greater diversity between evolved programs. 
\textsc{CheckMate}'s early stopping parameter $k$ (Sec.\,\ref{sec:heur-eval}) was set to $3$.
For all case studies, program evaluation criteria include built-in measures of code complexity and diversity, quantifying program length and (textual) differences. 
In addition, HCP's and CCP's criteria set contains the correctness, runtime, and memory usage scores. 
In contrast, E-DFJSP contains the quality and tardiness scores to guide the evolution more prominently towards programs that yield high-quality solutions.
To balance performance, efficiency, and cost ~\cite{novikov2025alphaevolve}, we used as LLMs GPT-5 for 60\% of queries and GPT-5-mini for the remaining 40\%.

\paragraph{Testing.} 
Each training run outputs a best program \ep, which was evaluated using \textsc{CheckMate} (standalone; without OpenEvolve) on the described test set (Sec.~\ref{sec:datasets}).
To illustrate the performance variability across the four training runs, we focus on \epl and \eph.
Here, \epl (\eph) denotes the evolved program that attains the lowest (highest) combined score on the test set, largely driven by the overall solving percentage.

\paragraph{Comparison with the baseline solvers and verifiers.}
To compare our evolved programs against state-of-the-art approaches, we executed top-performing solvers as standalone solving engines for each problem. 
For HCP and CCP, we employed \clingo with the original formal specifications (HCP~\cite{semmelrock26}; CCP~\cite{gebser2017sixth}), as it demonstrated strong solving performance in tests compared to a leading CP solver OR-Tools~\cite{perron_et_al:LIPIcs.CP.2023.3}. For {E-DFJSP}, we used CP-Optimizer, following the results reported in~\cite{da2022industrial}, together with a formal specification extending that of~\citeauthor{zuccato2025energy}~(\citeyear{zuccato2025energy}) in terms of a more general energy consumption formulation.

\subsection{Synopsis of the Best Evolved Programs}
The synopses of the best evolved programs were produced through a manual analysis conducted by the authors.
\paragraph{HCP.} % best_program_20260114_213701
HCP's \eph
begins by verifying available capacities by computing the minimum number of cabinets and rooms required under the ownership and capacity constraints and checks whether they exist. It proceeds greedily and deterministically, sorting things, cabinets and rooms and grouping them by owner, which enforces the ordering constraint. The program produces a correct configuration whenever sufficient resources exist, without exploring alternative assignments.

\paragraph{CCP.} % program 001211
CCP's \eph combines heuristic-guided search with constraint-aware pruning and bounded backtracking to construct correct configurations over a state space of partial color, bin, and area assignments. Instead of exhaustively enumerating possibilities, the program incrementally builds a complete assignment while discarding infeasible branches early based on capacity, path, and area constraints, and applying deterministic ordering for repeatability. The procedure is not purely greedy—rejected assignments may be revisited through limited backtracking—yet it avoids full search by steering decisions via deterministic heuristics. 

\paragraph{E-DFJSP.}
\label{sec:edfjsp-synopsis}
E-DFJSP's \eph consists of a greedy scheduler that first determines the set of eligible workers for each machine and, for every cut-machine pair, selects the best machine parameters by choosing the one with minimum processing time and, in case of ties, lower energy consumption. Jobs are ordered using the Earliest Due Date (EDD) heuristic to reduce total tardiness, and operations are scheduled respecting precedence constraints.
Per task, all feasible machine-worker-mode combinations are evaluated to minimize total completion time, processing energy, and resulting machine makespan.
The program maintains sorted calendars for machines, considering the interval from load start to unload completion, and for workers, who are required only during setup operations. Finally, it outputs a correct schedule together with the corresponding objective values.
\subsection{Results and Discussion}
\label{sec:results}

%%%%%%%%%
\begin{table}[t]
    \centering
    \setlength{\tabcolsep}{3pt}
\begin{tabular}{l|rrr|rrr|rrr}
\toprule
 Problem & \multicolumn{3}{c}{HCP} & \multicolumn{3}{c}{CCP} & \multicolumn{3}{c}{E-DFJSP}\\
  & \epl & \eph & \sol & \epl & \eph & \sol & \epl & \eph & \sol   \\
 \midrule
 Easy    & 100 & 100 & 100 & 63 & 93 & 100 & 100 & 100 & 100 \\
 Moderate    & 100 & 100 &  67 & 88 & 85 & 88 & 100 & 100 & 100\\
 Hard & 100 & 100 & 0 & 96 & 88 & 4 & 100 & 100 & 42\\
 \hline
Overall & 100 & 100 & 28 & 82 & 89 & 65 & 100 & 100 & 61\\
\bottomrule
\end{tabular}
% TODO EVENTUALLY SAME TABLE WITH AVG RUNTIME
\caption{Percentage of test instances solved by the evolved programs with the lowest and highest overall percentage (\epl and \eph) and by the state-of-the-art solver (\sol).}
    \label{table_results}
\end{table}

Tab.~\ref{table_results} summarizes the overall and difficulty-wise solving percentage for \epl, \eph, and the corresponding state-of-the-art solver (\sol) across the three selected case studies.
For HCP and E-DFJSP, both \epl and \eph solve all instances, whereas for CCP they solve $82\%$ and $89\%$, respectively. In contrast, \sols perform well on easy instances but degrade substantially on the moderate and hard ones, achieving $28\%$, $65\%$, and $61\%$ overall on HCP, CCP, and E-DFJSP, respectively.
This indicates that both \epl and \eph are competitive and consistently outperform traditional solvers on harder instances, underscoring the scalability of our approach.

\begin{figure*}[t] 
    %\flushleft       % aligns figure to the left
    \includegraphics[width=\textwidth]{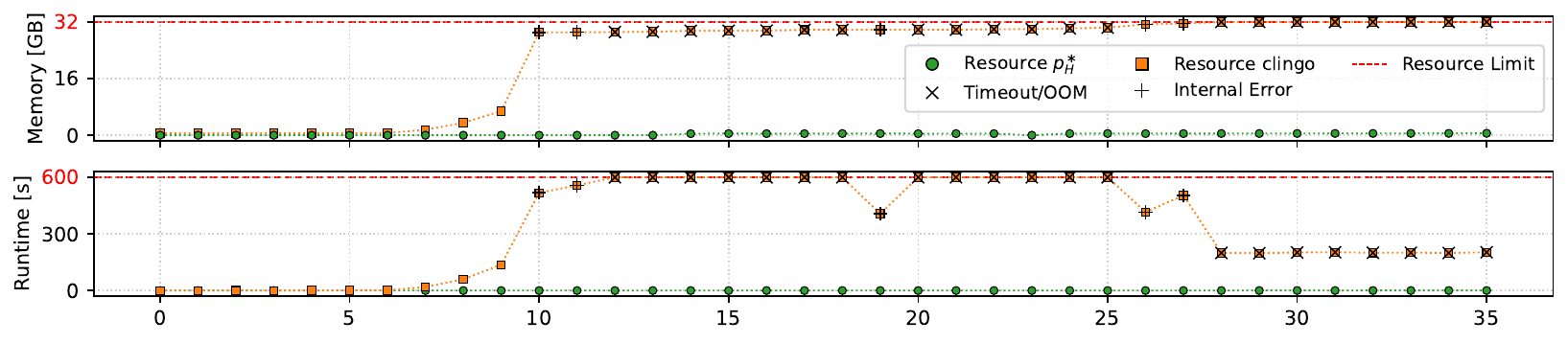}
    \caption{\textbf{HCP: Comparison between the evolved program (\eph) and \clingo} wrt.\ memory $[\mathit{GB}]$ (above) and runtime $[s]$ (below). Symbols $\times$, $+$ depict the reason why \clingo did not find a correct solution (OOM stands for out-of-memory). \eph is always correct.}
    \label{fig:HCP_cactus}
\end{figure*} 

\begin{figure*}[t] 
    %\flushleft       % aligns figure to the left
    \includegraphics[width=\textwidth]{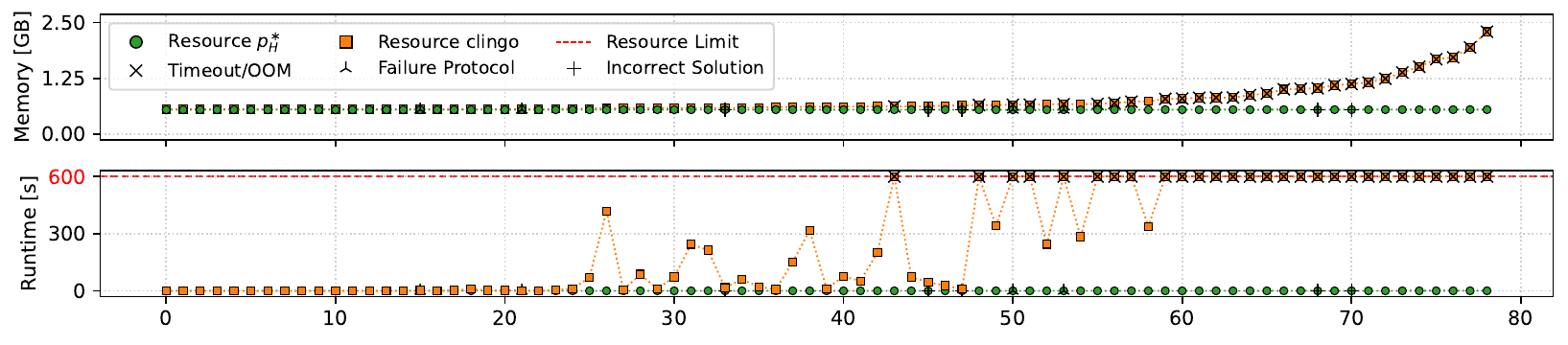}
    \caption{\textbf{CCP: Comparison between the evolved program (\eph) and \clingo} wrt.\ memory $[\mathit{GB}]$ (above) and runtime $[\mathit{s}]$ (below). Symbol $\times$ depicts the reason why \clingo did not find a correct solution (OOM stands for out-of-memory). The other symbols refer to \eph.}
    \label{fig:CCP_cactus}
\end{figure*} 

\begin{figure*}[t] 
    %\flushleft       % aligns figure to the left
    \includegraphics[width=\textwidth]{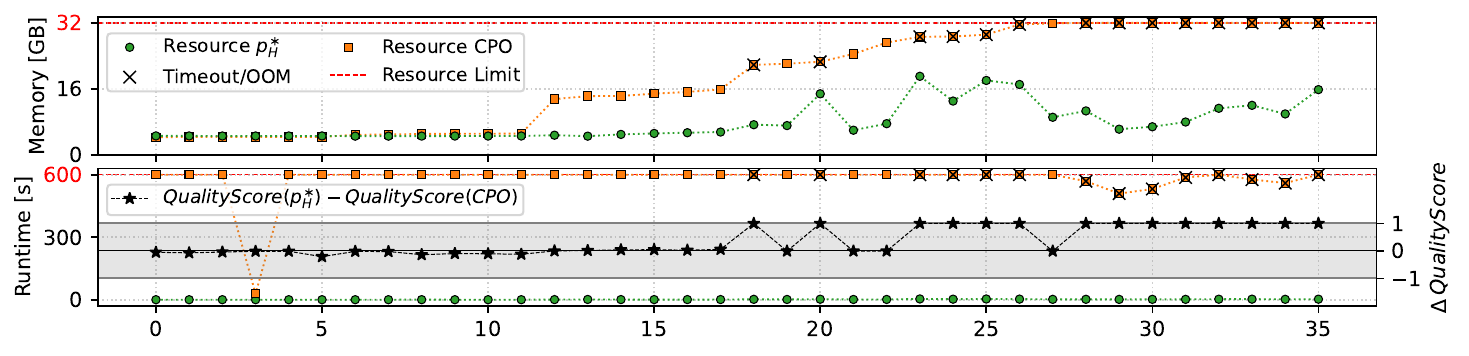}
    \caption{\textbf{E-DFJSP: Comparison between the evolved program (\eph) and CP Optimizer (\cpo)} wrt.\ memory $[\mathit{GB}]$ (above), runtime $[\mathit{s}]$ (below, primary y-axis), and the difference on their quality score (below, secondary y-axis, gray background). If the difference is positive then \eph is better than \cpo (when \cpo does not find a solution, its quality score is 0). Symbol $\times$ depicts the instances where \cpo did not find a correct solution (OOM stands for out-of-memory). \eph is always correct.}
    \label{fig:EDFJSP_cactus}
\end{figure*} 
The cactus plots depicted in Figs.~\ref{fig:HCP_cactus}, \ref{fig:CCP_cactus}, and \ref{fig:EDFJSP_cactus} compare correctness, memory usage (top panel), and runtime (bottom panel) between \eph and \sol for each problem. 
Instances are ordered by increasing memory usage of \sol. 

\paragraph{HCP.}
Fig.~\ref{fig:HCP_cactus} shows a clear performance gap on HCP. While both approaches handle the easy instances, \clingo's memory usage increases sharply at instance 11 and remains near saturation, leading to repeated timeouts (13 times), out-of-memory events (8 times), and internal errors (5 times) on all harder instances. In contrast, \eph maintains negligible memory (average: 281\,MB) and runtime (average: 0.08\,s) throughout and produces no incorrect solutions. Overall, \eph solves all instances with orders-of-magnitude lower resource usage, indicating superior scalability.
\paragraph{CCP.}
As can be seen from Fig.~\ref{fig:CCP_cactus}, \eph manifests consistently negligible runtimes while \clingo exhibits three different runtime behaviors.
More specifically, \clingo solves 28 instances within 10\,s and 23 instances between 10\,s and the 10\,min time limit, while 28 instances remain unsolved.
In contrast, \eph solves 89\% of the instances with consistently low memory and runtime, 546\,MB and 0.12\,s on average.
Given this high performance, which applies to all four best programs, even a parallel portfolio solver combining all of them is practical and achieves a 100\% solving rate across all training and test instances.
%This yields two key findings: \emph{(i)}~\eph solves more instances than the state-of-the-art solver, and \textit{(ii)}~it exhibits much lower resource usage.
This yields two key findings: \emph{(i)}~the best program \eph solves more instances \emph{and} exhibits a much lower resource usage than the state-of-the-art solver, and \emph{(ii)}~a portfolio of all four evolved programs can, to the best of our knowledge, \emph{for the first time}, successfully solve all CCP instances from \cite{Gebser2015SolvingCC}.
%This demonstrates that evolved programs can be superior to state-of-the-art approaches in terms of feasibility, robustness, computational efficiency, and scalability.
%This demonstrates that evolved programs can outperform a state-of-the-art solver on CCP in robustness and computational efficiency.  

\paragraph{E-DFJSP.}
% Fig.~\ref{fig:EDFJSP_cactus} compares \eph against CP Optimizer (\cpo) for E-DFJSP. The top plot reports memory consumption: for the twelve \textit{easy} and \textit{moderate} instances, both show similar values. 
% Thereafter, however, \eph consistently maintains a lower memory footprint (average: 7.89\,GB), whereas \cpo's memory usage escalates, leading to frequent resource errors. 
% Consequently, \cpo fails to solve 14 of the 18 \textit{hardest} instances. 
% The bottom plot exhibits runtime (primary y-axis) and solution quality (secondary y-axis). 
% \eph solves all instances with an average time of 1.26\,s, while \cpo finds the optimum only for the \textit{easiest} instance (10 jobs, 1 machine); in all other cases, it either fails or reaches the timeout.
% % , while \cpo stops only when it finds the optimum, i.e., only on the 
% The quality comparison (gray shaded area) demonstrates that, when both methods find a solution, the quality is generally comparable. Notably, \eph always produces a solution, while \cpo fails in $14$ instances.
% Overall, the state-of-the-art solver cannot solve most instances with at least 500~jobs in 10~minutes, while \eph always succeeds in less than 4\,s.
Fig.~\ref{fig:EDFJSP_cactus} compares \eph against CP Optimizer (\cpo) for E-DFJSP. With regard to the solving rate, 
%we see that 
\eph is successful 100\,\% of the time on the test instances, whereas \cpo fails to produce any solution on 14 of the 18 hardest ones due to 6 timeouts and 8 memory exhaustions.
Regarding memory consumption, the two approaches exhibit comparable behavior on the %twelve
\textit{easy} and \textit{moderate} instances; for the \textit{hard} ones, however, \cpo's space requirements escalate.
In terms of runtime, %for solved instances, 
\eph requires an average of 1.26\,s across all instances to find a solution, % (and stops, being a greedy algorithm Sec.~\ref{sec:edfjsp-synopsis}), 
whereas \cpo % fails %to produce any solution for 14 instances %(6 due to timeout) 
%and, for all other instances, it 
%does not produce a substantially better quality score than \eph despite consuming the entire time budget. % 
either fails to produce any solution %within the time limit 
or achieves, if at all, only marginally better quality scores 
%wrt.\ \eph 
despite using the entire time budget.
%, seeking optimality, hits the timeout for all but one instance. % proving optimality only for the easiest instance
% Concerning solution quality, whenever both methods produce a solution, their quality score (cf. Sec.~\ref{sec:edfjsp}) is comparable. 
%In cases where the scores differ, this is because \cpo does not produce any solution at all.
In fact, \cpo finds the optimum only for the easiest instance (10 jobs, 1 machine). 
Notably, for all instances with $200$ jobs or more, \eph outperforms \cpo on tardiness---the objective of highest priority---by 49\% on average.
Furthermore, the state-of-the-art solver cannot solve most instances with at least 500~jobs within 10\,min, while \eph always succeeds in less than 4\,s.

% \eph has a slightly worse average quality score (0.45 and 0.48, respectively), 
% \cpo average objective values are $\langle 38833, 184996, 666\rangle$ (tard[min], energy[kw], makespan[min] \ref{sec:casestudies}), while \eph's are $\langle 27450, 218065, 811 \rangle$ (filtering the instances solved by both). Therefore, as the optimization is lexicographic, \eph yields solutions with much higher quality in practice (at the price of additional energy and total time).

\subsubsection{Training and Checking Costs}
The cost in API fees was at most €\,20 per training run and below €\,200 overall.
Times per training run ranged from 1 to 16
hours, depending on the number of iterations and the efficiency of the generated programs. 
\textsc{CheckMate}'s average and, respectively, worst-case costs for checking the correctness of outputs of best evolved programs \eph amounted to 3.01\,s / 0.11\,s / 1.93\,s and 18.28\,s / 0.32\,s / 5.13\,s per test instance for HCP / CCP / E-DFJSP.

%%%%%%%%%%%%%%%%%%%%%%%%%%%%%%%%%%%%%%%%%%%%%%%%%%%%%%%%%%%%%%%%%%%%%%%%%%%%
\section{Related Work}
\label{sec:related}

\paragraph{Combinatorial and optimization problems.}
Modern symbolic AI methods utilize declarative formalisms, such as Answer Set Programming  \cite{DBLP:series/synthesis/2012Gebser}, Constraint Programming  \cite{DBLP:reference/fai/RossiBW08}, or Mixed Integer Programming \cite{wolsey2020integer}, 
to address complex combinatorial and optimization problems. These approaches decouple problem specification from search algorithms, enabling domain-independent solvers to find optimal or near-optimal solutions using advanced techniques like conflict-driven clause learning and constraint propagation.
However, large industrial problem instances significantly reduce solver performance in practice  \cite{DBLP:journals/ki/FalknerFSTT18,schlenkrich2023solving}. To improve performance, research has focused on three directions \cite{DBLP:conf/ijcai/KotaryFHW21}: \emph{(i)}~developing domain-specific heuristics, \emph{(ii)}~improving problem encodings, and \emph{(iii)}~configuring existing or learning new problem-specific algorithms. While designing any of these approaches requires significant domain expertise and manual effort, it has been observed that machine learning (ML) methods can simplify this challenge in many industrial cases where problem instances share similar patterns.

For the first direction, various ML techniques have been proposed to learn effective heuristics from, e.g., problem instances or solving traces, using supervised or reinforcement learning methods \cite{bengio2021machine,lodi2017learning}. Injected into a solver, learned heuristics can significantly enhance its performance in the target domain. 
In the second direction, ML has been used to generate or optimize problem encodings, e.g., by adding symmetry-breaking or implied constraints \cite{DBLP:conf/aaai/TarzariolGSL23,DBLP:journals/tplp/TaupeWF20}, or finding problem decompositions \cite{DBLP:journals/jair/CappartGLPT25}. 
In the third direction, ML approaches were first used to develop portfolio solvers capable of automatically configuring existing solvers for specific problem instances \cite{DBLP:series/lncs/Kotthoff16}. More recently, deep learning methods have been proposed to solve combinatorial problems in an end-to-end fashion \cite{DBLP:conf/ijcai/KotaryFHW21}, learning to predict (approximate) solutions directly without invoking solvers at inference time.

Our approach can roughly be classified in the third direction, as we aim to automatically generate problem-specific solving algorithms. Unlike prior work that relies on deep learning models, we employ evolutionary computation combined with LLMs to synthesize code. 
This enables us to generate interpretable and verifiable algorithms, rendering our approach more flexible without compromising performance across different combinatorial and optimization problems.

\paragraph{Code generation for combinatorial optimization.} 
Code evolution with LLMs has recently emerged as a particularly relevant research topic in combinatorial optimization. Early frameworks, such as FunSearch \cite{Romera-Paredes2024}, focused on evolving heuristics for Cap Set and Bin Packing problems. Subsequent works, such as Evolution of Heuristics \cite{liu2024evolutionheuristicsefficientautomatic}, expanded the scope to a wider range of problems, such as online bin-packing, traveling salesman, and flow shop scheduling.
Furthermore, AlphaEvolve \cite{novikov2025alphaevolve} and its open-source implementations, such as OpenEvolve \cite{openevolve}, ShinkaEvolve \cite{shinkaevolve}, DeepEvolve \cite{deepevolve}, and CodeEvolve \cite{2026codeevolveopensourceevolutionary}, have applied code evolution to a variety of domains, focusing on the previously mentioned problems, as well as matrix multiplication, the minimum overlap problem, and kissing numbers in 11 dimensions.
The aforementioned approaches neither explicitly mention formal verification nor utilize software testing, but rely on handcrafted solution checkers.
More recently, the authors of \cite{georgiev2025mathematicalexplorationdiscoveryscale} applied AlphaEvolve to 67 mathematical problems, %across various domains, including combinatorics, mathematical analysis, geometry, and number theory
and, in a few cases, verified the program-generated solutions using AlphaProof~\cite{hubert_olympiad-level_2026} and Lean~\cite{deMoura2015lean}.
%, an interactive theorem prover.
%
Notably, SATLUTION \cite{yu2025autonomouscodeevolutionmeets} evolves entire code repositories to produce variants of SAT solvers, which operate at the propositional level.
Differently, \textsc{CheckMate} employs first-order specifications to verify the correctness of candidate solutions for the given problem instances.
Therefore, \textsc{CheckMate} addresses an open point in prior work by providing an automated declarative verification approach that guarantees the correctness of returned solutions.
%Therefore, \textsc{CheckMate} addresses one of the open points in prior research related to the automatic verification of solutions. 
%In particular, the aforementioned approaches either do not explicitly mention verification or utilize software testing and human-based verification, as in the case of AlphaEvolve. 
%\textsc{CheckMate} suggests a common verification approach that utilizes declarative solving techniques, thereby guaranteeing the correctness of the solutions returned.

%%%%%%%%%%%%%%%%%%%%%%%%%%%%%%%%%%%%%%%%%%%%%%%%%%%%%%%%%%%%%%%%%%%%%%%%%%%%
\section{Conclusions}
\label{sec:conclusion}

The OpenEvolve+\textsc{CheckMate} framework showed the great potential to automatically synthesize problem-solving algorithms implemented in Python.
The system needs no information on \emph{how} to construct solutions. It relies only on \emph{what} defines a correct (optimal) solution and a set of representative instances.
%
%Although 
The obtained programs %are incomplete solvers, just as most of the modern approaches relying on ML to combinatorial optimization, 
% they 
can efficiently tackle large and challenging instances from the practical use cases of Siemens and voestalpine that are currently out of reach for state-of-the-art solvers.
% Moreover, by using \textsc{CheckMate} as a verifier, we ensure the correctness of the output of the synthesized programs.
%
Our analysis demonstrates that the generated algorithms can successfully solve difficult real-world \emph{(a)}~combinatorial and \emph{(b)}~optimization problems such as configuration and scheduling, thereby addressing \ref{rq1:solving}. The evolved programs significantly outperform state-of-the-art solvers on hard instances, addressing \ref{rq2:vs-sota}. Finally, the results highlight strong scalability: both in terms of increasing problem size, as tested on the HCP and E-DFJSP, and in terms of increasing problem hardness, as examined with regard to the CCP. This addresses \ref{rq3:scalability} and confirms the practical viability of our approach.

Future research should explore several promising directions to further evaluate the approach and enhance its applicability and robustness.
For example, by using \textsc{CheckMate} to verify the outputs of synthesized programs, we ensure solution correctness per instance, while establishing their overall correctness remains important future work.
Moreover, we will conduct systematic ablation studies to quantify the contribution of every \textsc{CheckMate} component.

\begin{comment}
    
extending the framework to support a broader range of problem domains, including those with dynamic or stochastic constraints, could significantly expand its industrial relevance. Second, integrating more advanced feedback mechanisms, such as minimal unsatisfiable subsets (MUS) or explainable artifacts, could improve the interpretability and refinement of evolved programs. 
%Third, investigating hybrid approaches that combine code evolution with other ML, such as reinforcement learning, may yield even more powerful solvers. 
Finally, addressing challenges related to generalization, such as evolving programs that adapt to unseen structures in problem instances, remains an important challenge for future work. These advancements could further establish code evolution as a cornerstone methodology for solving complex combinatorial optimization problems.
\end{comment}

%\appendix
% optional, for technical details. Note: counts for the page limit!

%\section*{Ethical Statement}
% optional

\section*{Acknowledgments}
% only for camera ready
This research was funded in whole or in part by the Austrian Science Fund (FWF) 10.55776/COE12 and the Austrian Research Promotion Agency (FFG) FO999910235 (SAELING) and 930480ATRIA (ATRIA). % This research was also funded in part by the Austrian Science Fund (FWF) 10.55776/COE12.

\bibliographystyle{named}
\bibliography{ijcai26}

\end{document}